\documentclass[conference]{IEEEtran}
\IEEEoverridecommandlockouts
\usepackage{cite}
\usepackage{amsmath,amssymb,amsfonts}
\usepackage{algorithmic}
\usepackage{array} 
\usepackage{multirow}
\usepackage{booktabs}
\usepackage{url}
\usepackage{graphicx}
\usepackage{textcomp}
\usepackage{xcolor}
\def\BibTeX{{\rm B\kern-.05em{\sc i\kern-.025em b}\kern-.08em
    T\kern-.1667em\lower.7ex\hbox{E}\kern-.125emX}}
\begin{document}

\title{CountDiffusion: Text-to-Image Synthesis with Training-Free Counting-Guidance Diffusion}

\author{
    Yanyu Li, Pencheng Wan, Liang Han, Yaowei Wang, Liqiang Nie, Min Zhang \\
    Department of Computer Science and Technology, Harbin Institute of Technology, Shenzhen, China 
}

\maketitle

\begin{abstract}
Stable Diffusion has advanced text-to-image synthesis, but training models to generate images with accurate object quantity is still difficult due to the high computational cost and the challenge of teaching models the abstract concept of quantity. In this paper, we propose CountDiffusion, a training-free framework aiming at generating images with correct object quantity from textual descriptions. CountDiffusion consists of two stages. In the first stage, an intermediate denoising result is generated by the diffusion model to predict the final synthesized image with one-step denoising, and a counting model is used to count the number of objects in this image. In the second stage, a correction module is used to correct the object quantity by changing the attention map of the object with universal guidance. The proposed CountDiffusion can be plugged into any diffusion-based text-to-image (T2I) generation models without further training. Experiment results demonstrate the superiority of our proposed CountDiffusion, which improves the accurate object quantity generation ability of T2I models by a large margin. 
\end{abstract}


\section{Introduction}
\label{sec:intro}
\begin{figure}[htbp]
\centerline{\includegraphics[width=1\linewidth]{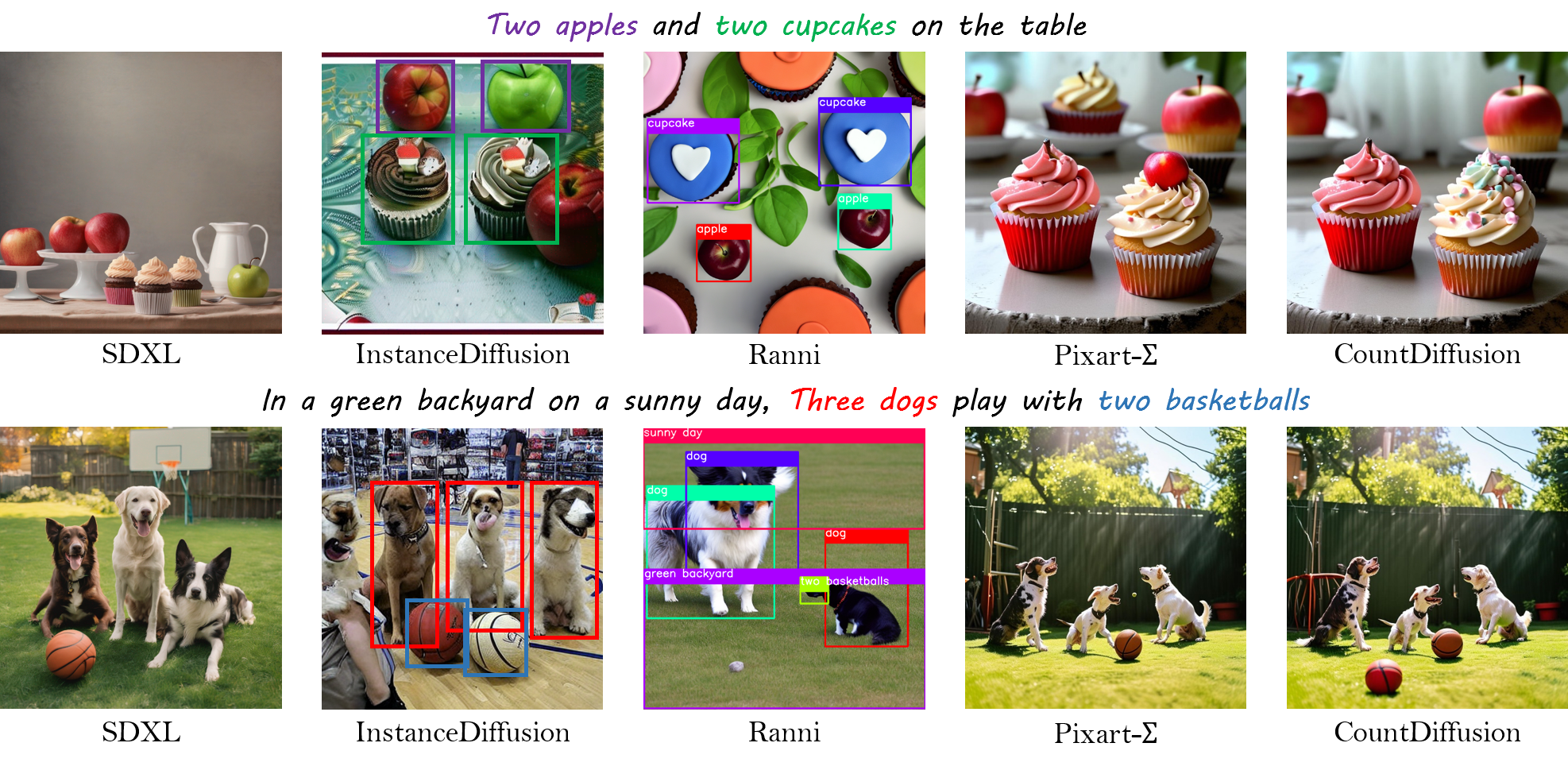}}
\caption{Samples generated by SDXL, Pixart-$\Sigma$, Ranni, InstanceDiffusion, and CountDiffusion (Ours). Existing models struggle with generating images with correct objects counts. Even when provided with bounding boxes, the models may still generate target objects in the background. Furthermore, Ranni, which combines a LLM for T2I generation, is limited by the language model and is still highly likely to generate incorrect objects counts. CountDiffusion can correct the object quantity based on the generation results of the baseline model, $i.e.$, Pixart-$\Sigma$.}
 \label{figure1}
\end{figure}

    When you are trying to describe a scene to your friend, what is your preferred choice, language or vision? Image is definitely a more straightforward and simpler way. Thanks to the text-to-image (T2I) synthesis, which aims at generating photo-realistic images based on textual descriptions, we can describe a scene with image easily. Benefiting from the recent advances in image generation models such as stable diffusion \cite{SDXL}, Midjourney\cite{Midjourney}, DALL-E \cite{dalle}, we now can obtain promising synthesized images in terms of high fidelity and diversity with some T2I synthesis models \cite{DALLE3, Imagen}. However, despite the powerful image generation ability, T2I synthesis models still struggle with understanding complex and abstract languages, and as a result, these models suffer from translating a textual description into a precisely corresponding image. For example, it is challenging for all the current T2I synthesis models, either open sourced models such as SDXL, ranni\cite{feng2024ranni} or proprietary models such as Midjourney, to generate objects with specified quantity in an image, as shown in Figure \ref{figure1}.

      How can we obtain generated images with accurate object quantities with T2I synthesis models? To solve this problem, researchers propose to leverage additional guidance such as poses, masks\cite{scenecomposer, dense}, bounding boxes\cite{boxdiff, layoutllm}, depth\cite{SDXL}, keypoints\cite{instancediffusion}, scribbles and canny edge\cite{ControlNet}, etc. along with textual descriptions to improve the accurate object quantity generation ability of T2I synthesis models,  resulting in a series of models such as ControlNet \cite{ControlNet}, Detector-Guidance \cite{detector}, ALDM \cite{li2024adversarial}, etc. However, these methods that require additional guidance information are usually user-unfriendly, especially when a non-expert user is asked to provide object masks or keypoints. Qu et al. \cite{layoutllm} take advantage of large language models (LLMs) to alleviate this problem. They first use a LLM to generate a layout with object bounding boxes based on the textual descriptions, and then generate images from the layout with an image generation model. InstanceDiffusion \cite{instancediffusion} further simplifies and unifies different guidance such as masks, scribbles, bounding boxes, etc., into points, allowing users to specify object quantity information with points. Unfortunately, these methods still struggle with generating images with correct objects counts, as shown in Figure \ref{figure1}. Besides, training these models is also prerequisite in order to involve the additional guidance information into the image generation process, which is expensive and time-consuming because of the huge number of parameters in the T2I synthesis models.
    
    To address these challenges, we introduce CountDiffusion in this work, a training-free approach that injects counting into diffusion-based T2I models to enable them to rectify object quantity during the image generation process, $i.e.$, the denoising process of the diffusion models. The basic idea of the proposed CountDiffusion is to obtain an intermediate generated image, which is used to guide the final generation by counting the object quantity in this intermediate generation. Precisely, an intermediate generation result is first sampled in a certain denoising step of a diffusion-based T2I model, which is used to predict a final generation result by one-step denoising. Then, a counting tool, $e.g.$, Grounded SAM\cite{SAM}, is applied on the predicted final generation result to count the number of generated objects, and determine some areas to add or remove objects if count number is inconsistent with the textual description. Lastly, an image is synthesized  with the step-by-step denoising of the diffusion model and guided by the counting information, which is regarded as the final synthesized result.

    The proposed CountDiffusion is training-free, and can be easily generalized to all diffusion-based T2I synthesis models. To support future work in this field, we also build a dataset with LLMs. The evaluation results on the built dataset and a public dataset demonstrate the superiority of the proposed approach, which can significantly improve the accuracy of the generated object quantity.

    The main contributions of this paper can be summarized as follows:
    
\begin{itemize}
  \item We propose a training-free framework, CountDiffusion, to guide diffusion models to generate images with accurate object quantity based on textual descriptions by combining the diffusion-based T2I synthesis model with an object counting model. The training-free property saves the proposed model from training with a large number of data, which is costly and time-consuming. Furthermore, the proposed approach can be generalized to all diffusion-based T2I synthesis models.
  \item A multi-loss universal guidance approach is designed which extends support from a single loss to multiple losses. This addresses the issue of competition among multiple losses, greatly enhancing the stability of universal guidance and expanding its applicability.
  \item To promote the future research on this topic, a dataset is built with LLMs, which consists of textual descriptions for both single-class objects and multi-class objects. Extensive experiments are conducted on the built dataset and a public dataset, which demonstrate the effectiveness of the proposed method on accurate T2I object quantity generation.
\end{itemize}

\section{Related Work}
  \begin{figure*}[htbp]
      \centering
      \includegraphics[width=1\linewidth]{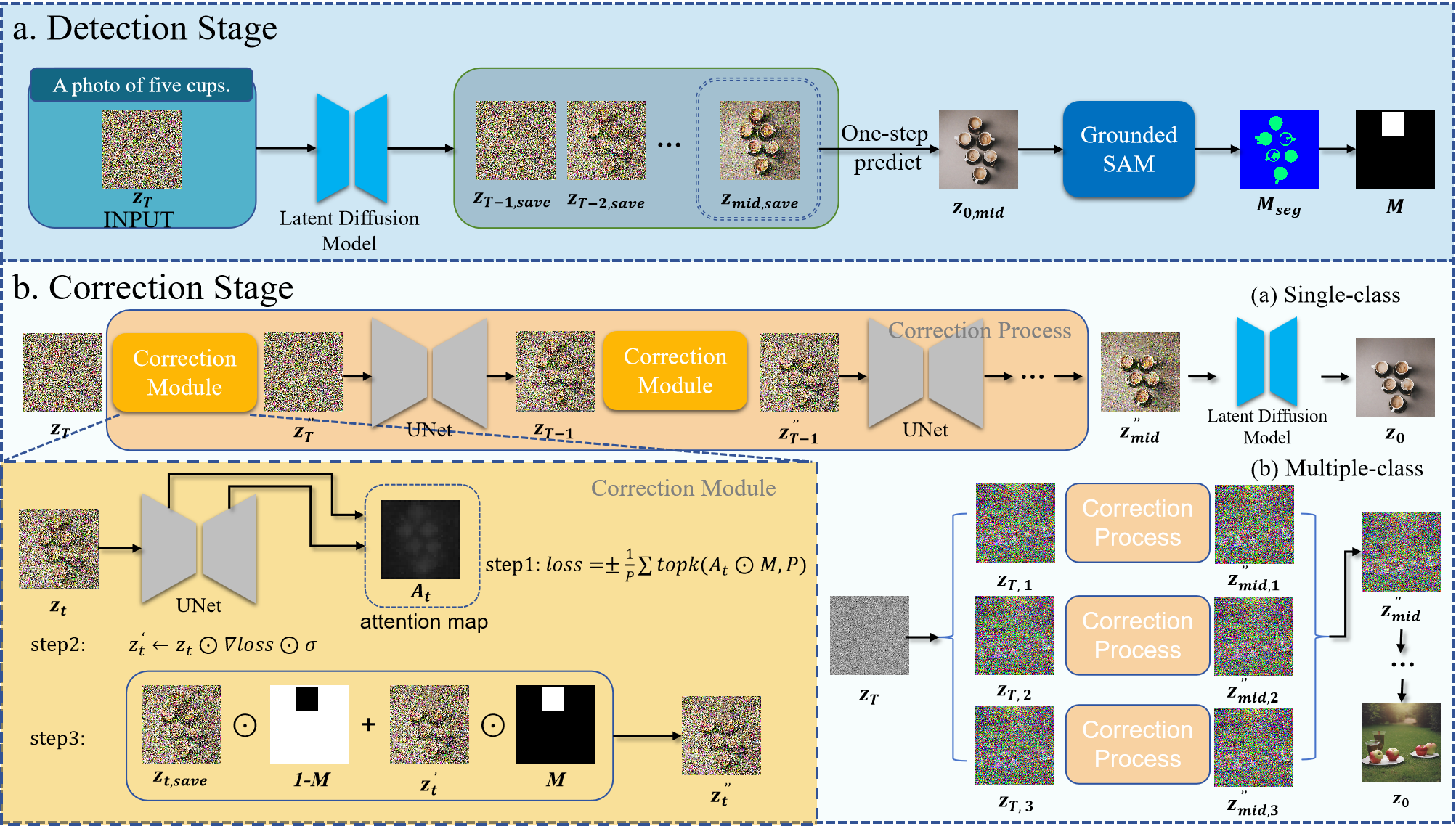}
      \caption{Pipeline of the proposed CountDiffusion. CountDiffusion consists of a detection stage and a correction stage. In the detection stage, an intermediate denoising result generated by the diffusion-based T2I model is used to predict the final synthesized image with a one-step denoising. Then, a counting model, e.g., Grounded SAM, is used to get the quantity and segmentation information of the objects in this image. In the correction stage, a correction module is used to rectify the object quantity by modifying the latent feature of the synthesized image with universal guidance. Different correction strategy for single-class objects and multi-class objects.}
      \label{figure2}
  \end{figure*}

  \textbf{Diffusion Models:} Diffusion models have become the mainstream approach in T2I synthesis due to their ability of generating images with high fidelity and diversity\cite{denoising, SDXL, High-resolution}. Diffusion models consist of a forward process that adds noise and a backward process that remove noise, both are step-by-step. After training, diffusion models can progressively convert a Gaussian noise image into a realistic image by denoising in each step of the backward process. Currently, there are many variants of diffusion models. The denoising diffusion probabilistic model (DDPM) \cite{denoising} learns to invert a parameterized Markovian process of noise adding. Various guidance information are involved into the image generation process by inserting cross attention layers in the diffusion models to achieve multimodal image synthesis tasks such as text-to-image synthesis and image-guided image generation\cite{choi2021ilvr, dhariwal2021diffusion, meng2021sdedit}. Furthermore, in order to accelerate the denoising process, the Latent Diffusion Model (LDM) \cite{deep} uses an AutoEncoder to project images into the latent space, which greatly reduces computational cost. Due to their strong performance, diffusion models are now also being employed in the domains of video generation\cite{ModelScope, magictime, champ} and 3D synthesis\cite{anciukevivcius2023renderdiffusion, qian2023magic123}.

  \textbf{Text-to-Image models:} Because of the powerful image generation ability, GAN-based models\cite{reed2016generative, reed2016learning} have been the mainstream approaches in T2I generation and have made great progress since 2016. However, GANs face the model collapse problem. In 2015, diffusion model\cite{deep} was introduced. Unfortunately, the original diffusion model has a huge computational cost and the inference time is quite long. It was not until 2021 that Dhariwal et al.\cite{dhariwal2021diffusion} proved that diffusion theory supports text guidance and proposed classifier guidance diffusion. It trained an additional classifier to guide the image generation. However, at that time, the diffusion model only supported input of object categories. Fortunately, in late 2021, Liu et al.\cite{liu2023more} expanded the classifier. Since then, the diffusion model has been able to achieve controlled generation of free text. However, the drawback is that while training a diffusion model, a classifier also needs to be trained. This not only increases the difficulty of training but also increases the cost of inference. In 2022, Ho et al.\cite{ho2022classifier} proposed Classifier-Free Diffusion Guidance, theoretically proving that the diffusion model does not require a classifier. Since then, the diffusion model has firmly established its position in the field of T2I generation. 

  \textbf{Universal guidance:} Universal guidance\cite{universal} is a guidance algorithm that augments the image sampling method of a diffusion model to include guidance from an off-the-shelf auxiliary network.  It introduces a guidance loss during the backward process of the diffusion model, where the loss is back-propagated a few times at each step to achieve guided image generation. Compared to training-based algorithms, this algorithm only adds a small amount of inference time but achieves comparable results in fine-tuning and continued training. It performs well in various domains such as style transfer\cite{style}, conditional generation\cite{boxdiff}, and image editing\cite{prompt2prompt}.

\section{Preliminaries: Stable Diffusion}
  The stable diffusion model consists of three main components: an autoencoder with an encoder $\mathcal{E}$ and a decoder $\mathcal{D}$, and a denoiser $\epsilon_\theta$. Given an image $x$, the encoder $\mathcal{E}$ maps it to the latent space, $i.e.$, $z = \mathcal{E}(x)$, and the decoder reconstructs the image from the latent space, $i.e.$, $x = \mathcal{D}(z)$. The denoiser is used repeatedly $T$ times to predict the next latent representation with less noise:
  \begin{equation}
      \epsilon_t = \epsilon_\theta(z_t, t),
      \label{eq-3}
  \end{equation}
  \begin{equation}
      z_{t-1} = \mathcal{M}(z_t, t) = \frac{z_t-(1-\sqrt{1-\overline{\alpha_t}}\epsilon_t)}{\sqrt{\overline{\alpha_t}}},
      \label{eq-1}
  \end{equation}
  where $t\in{T}$ represents the current time step, $\epsilon_t$ represents noise predicted by denoiser, and $\lbrace\overline{\alpha}\rbrace_{t=1}^T$ are hyperparameters. And finally, the decoder decodes $z_0$ to the clean image, $i.e.$, $\overline{x} = \mathcal{D}(z_0)$. In addition, we can directly predict the final result from any intermediate time step.:
  \begin{equation}
      z_0 = \mathcal{P}(z_t, t) = \frac{z_t-(1-\sqrt{1-\alpha_t^{'}}\epsilon_t)}{\sqrt{\alpha_t^{'}}},
      \label{eq-2}
  \end{equation}
  where $\lbrace\alpha^{'}\rbrace_{t=1}^T$ are hyperparameters. In the diffusion model, the assumption that the reverse steps are a Markov process relies on the condition that the denoising strength is sufficiently small at each step. Although one-step prediction violates this assumption, the generated images still provide enough positional and quantity information. Thanks to Dhariwal et al. \cite{dhariwal2021diffusion} and Liu et al. \cite{liu2023more}, the stable diffusion model is now supporting text conditioning, making accurate object quantity generation possible. With the addition of an extra text encoder $\tau$ and text input $Y$, obtained from Eq.\ref{eq-3}, Eq.\ref{eq-1} and Eq.\ref{eq-2}, we have:
  \begin{equation}
      y = \tau(Y),
  \end{equation}
  \begin{equation}
      \epsilon_t = \epsilon_\theta(z_t, y, t),
  \end{equation}
  \begin{equation}
      z_{t-1} = \mathcal{M}(z_t, y, t) = \frac{z_t-(1-\sqrt{1-\overline{\alpha_t}}\epsilon_t)}{\sqrt{\overline{\alpha_t}}},
      \label{eq1}
  \end{equation}
  \begin{equation}
      z_0 = \mathcal{P}(z_t, y, t) = \frac{z_t-(1-\sqrt{1-\alpha_t^{'}}\epsilon_t)}{\sqrt{\alpha_t^{'}}}.
      \label{eq2}
  \end{equation}
\section{Method}

  The proposed CountDiffusion can be applied to any diffusion-based T2I model without additional training. It consists of two stages as shown in Fig. \ref{figure2}, detection stage and correction stage. In the detection stage, we count the objects in the image and obtain the mask of the region to be rectified (section \ref{Selection Module}). In the correction stage, the correction module rectifies the object quantity (section \ref{Correction Module}). 

  \subsection{Detection Stage}
  \label{Selection Module}
\begin{figure}[!t]
      \centering
      \includegraphics[width=1\linewidth]{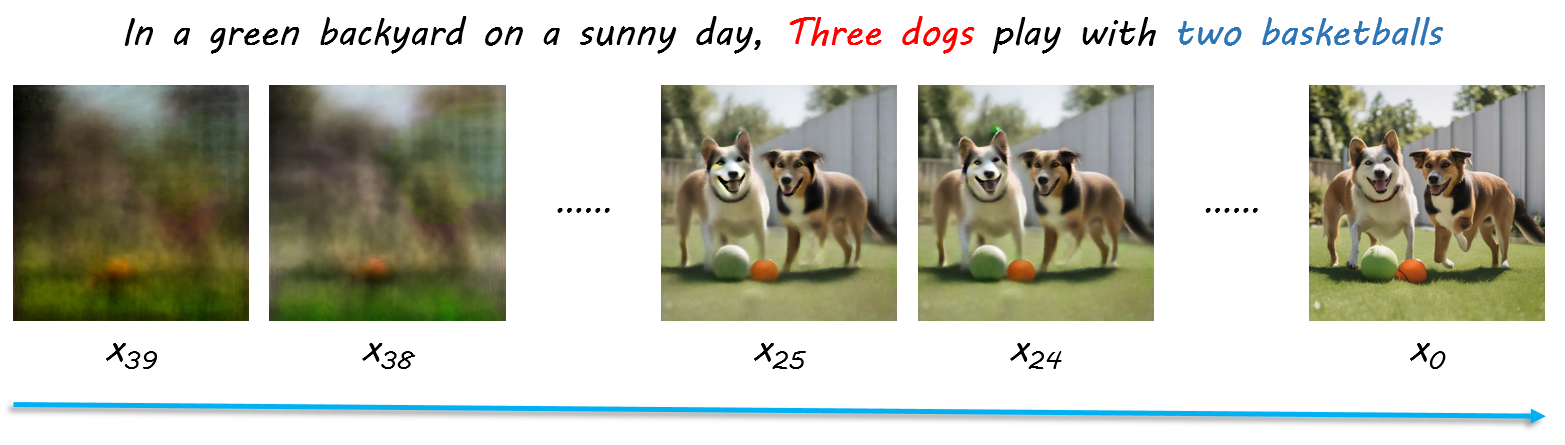}
      \caption{The synthesized images from intermediate denoising results of different denoising steps with one-step denoising. Total denoising step $T=40$ in this case. After approximately 15 denoising steps (i.e., $t_{mid}=25$), the image generated with one-step denoising is already able to provide sufficient object quantity and position information.}
      \label{figure3}
  \end{figure}

  The detection stage adopts a diffusion model to perform denoising from the initial step $T$ to the intermediate step $t_{mid}$ guided by the input textual description, while recording latent features $\{ z_{t,\text{save}} \mid t \in \{mid, mid+1, ..., T \}$. Then, $z_{mid, save}$ along with the decoded text information $y$ are used to predict the final denoising result $z_{0, mid}$ with a one-step denoising, and the decoder $\mathcal{D}$ is used to convert $z_{0, mid}$ back to the image space and get a synthesized image $x_{0, mid}$:
  \begin{equation}
      z_{0, mid} = \mathcal{P}(z_{mid, save}, y, t),
  \end{equation}
  \begin{equation}
      x_{0, mid} = \mathcal{D}(z_{0, mid}).
  \end{equation}
    Though predicting $x_{0, mid}$ in a single step violates the Markov chain assumption of the diffusion model and leads to poor image quality, it still accurately capture the object quantity information and object positions as shown in Fig. \ref{figure3}. Finally, Grounded SAM ($\mathcal{SAM}$) is adopted to segment and recognize each object in the predefined classes in image $x_{0, mid}$, with which we can obtain the quantity of each class of objects and their corresponding segmentation masks $M_{seg}$, $i.e.$, 
  \begin{equation}
      M_{seg} = \mathcal{SAM}(x_{0, mid}, tags), M_{seg} \in \mathbb{N}^{W, H},
  \end{equation}
where $tags$ refers to the objects that need to be synthesized. These tags can be provided by the user or determined by LLMs using the input textual description. $W$ and $H$ denote the width and height of the synthesized image, respectively. By comparing the quantity of the synthesized objects with the object quantity in the textual description, we can obtain the correction region mask $M$ from $M_{seg}$. Then, we eliminate objects by reducing the attention map values within the correction region mask $M$ if more objects are generated than required. Otherwise, objects can be added by increasing the attention map values in the non-overlapping regions of the image. 
\subsection{Correction Stage}
\label{Correction Module}

  In the correction stage, the input Gaussian noise image is the same as the one in the dectection stage. During the correction process, that is, from the initial step $T$ to the intermediate step $t_{mid}$, correction module is used to correct the object quantity by modifying the attention values of the attention map. The remaining denosing steps keep the same with DDIM. 
  
  Specifically, when computing $z_{t-1}$ from $z_{t}$, attention map $A_t$ is simultaneously returned: 
  \begin{equation}
      z_{t-1}, A_t = \mathcal{M}(z_t, y, t) , \quad t \in {1, 2, ..., T}. 
      \label{eq3}
  \end{equation}
  
  The loss function is designed based on the returned attention maps $A_t$ and the selected object masks obtained from the sampling stage, and new $z_t^{''}$ is generated in a backward manner using universal guidance, 
  \begin{equation}
    loss = \pm \frac{1}{P}\sum{\mathbf{topk}(A_{t} \odot M, P)}, t \in {mid, mid+1, ..., T},
    \label{eq4}
  \end{equation}
  \begin{equation}
      z_t^{'} \leftarrow z_t \odot \nabla loss \odot \sigma, \quad t \in {mid, mid+1, ..., T},
  \end{equation}
  \begin{equation}
      z_t^{''} = z_t^{'} \odot M + z_{t, save} \odot (1- M), \quad t \in {mid, mid+1, ..., T},
  \end{equation}
  where $\mathbf{topk}(A_{t} \odot M, P)$ means that P$\%$ elements with the largest attention values would be selected, $\odot$ denotes the element-wise multiplication of two matrices, and $\sigma$ serves as a control scale for the intensity. The variable $z_{t, save}$ means the intermediate results saved during the sampling stage, which helps preserve the background (i.e., region outside the $M$) unchanged. In other words, the loss calculates the average of the top P$\%$ largest attention values of the correction region (inside the $M$). Note that when the number of generated objects is large than the requirement, we apply a positive loss function. Otherwise, a negative loss function is applied. Besides, we find that applying Gaussian smooth on the attention map before calculating the loss can greatly improve the image quality and the object quantity correction accuracy. The universal guidance is only applied from the initial step $T$ to the intermediate step $t_{mid}$ in order to guarantee the high fidelity of the synthesized image.

  \subsection{Multi-class Object Correction Strategy}
  \label{Multi-class Objects Correction Strategy}
  When correcting the quantities of multi-class objects on a single attention map, the correction losses of objects in different classes may affect each other and result in unstable loss optimization. Therefore, we design a distribution correction strategy to minimize this interaction. Specifically, from the initial step $T$ to the intermediate step $t_{mid}$, we apply the same initial Gaussian noise to objects in each class separately, and a single-class correction process is performed individually for each class. Then, taking the average of the latent features $z_{mid, i}$ obtained by correcting quantity for each object class in step $t_{mid}$ as the universal guidance result of the multi-class objects. Mathematically, 
  \begin{equation}
      z_{mid} = \frac{1}{n} \sum_{i=1}^{n} z_{mid, i}, \quad i \in {1,2, .., n},
  \end{equation}
where $n$ is the number of object classes of which quantities need to be
rectified.

\section{Experiments}
\subsection{Evaluation Dataset}

\begin{table*}[htbp]
\caption{The comparison results of SDXL, Pixart-$\Sigma$, CountGen and CountDiffusion in terms of Acc.($\%$), MAE, CLIP-score and ImageReward. (CountGen, based on SDXL, is limited to generate images with single-class objects.)}
\centering
\begin{tabular}{ccccccccccccc}
\toprule
\multirow{3}{*}{\textbf{Model}} &  \multicolumn{8}{c}{\textbf{Single-class}} &  \multicolumn{4}{c}{\textbf{Multi-class}} \\
\cline{2-13}
 &\multicolumn{4}{c}{\textbf{CoCoCount}} & \multicolumn{4}{c}{\textbf{GPTSingleCount}} & \multicolumn{4}{c}{\textbf{GPTMultiCount}} \\
\cline{2-13}
 & Acc.$\uparrow$ & MAE$\downarrow$ & CLIP-score$\uparrow$ & IR$\uparrow$ & Acc.$\uparrow$ & MAE$\downarrow$ & CLIP-score$\uparrow$ & IR$\uparrow$ & Acc.$\uparrow$ & MAE$\downarrow$ & CLIP-score$\uparrow$ & IR$\uparrow$ \\
\midrule
SDXL & \multicolumn{1}{|c}{34} & 2.33 & 32.5038 & 0.89 & \multicolumn{1}{|c}{21} & 4.16 & \textbf{31.7788} & 0.62 & \multicolumn{1}{|c}{5} & 2.10 & \textbf{32.9595}&  0.71\\
CountGen & \multicolumn{1}{|c}{51} & 1.28 & 32.0375 & 0.94 & \multicolumn{1}{|c}{35} & 2.49 & 31.3481 & 0.45 & \multicolumn{1}{|c}{-} & - & - & - \\
CountDiffusion (SDXL) & \multicolumn{1}{|c}{\textbf{59}} & \textbf{0.90} & \textbf{33.1402} & \textbf{1.02} & \multicolumn{1}{|c}{\textbf{45}} & \textbf{1.78} & 31.3483 & \textbf{0.69} & \multicolumn{1}{|c}{\textbf{31}} & \textbf{1.38}& 32.7549& \textbf{0.82} \\
\midrule
Pixart-$\Sigma$ & \multicolumn{1}{|c}{40} & 1.33 & 31.9596 & 0.96 & \multicolumn{1}{|c}{22} & 2.77 & 31.1993 & 1.02 & \multicolumn{1}{|c}{23} & 1.35 & 33.0638& 1.24\\
CountDiffusion (Pixart-$\Sigma$) & \multicolumn{1}{|c}{\textbf{60}} & \textbf{0.89} & \textbf{32.0006} & \textbf{1.21} & \multicolumn{1}{|c}{\textbf{37}} & \textbf{2.22} & \textbf{31.2482} & \textbf{1.02} & \multicolumn{1}{|c}{\textbf{43}} & \textbf{0.86}& \textbf{33.3786}& \textbf{1.30} \\
\bottomrule
\end{tabular}
\label{table1}
\end{table*}

We evaluate our method on three datasets: the public CoCoCount dataset and two self-constructed datasets, GPTSingleCount and GPTMultiCount. Both self-constructed datasets comprise the same 25 randomly selected object categories, including: 'red apple', 'car', 'wooden chair', 'golden retriever dog', 'glass table', 'pine tree', 'sunflower', 'parakeet bird', 'office worker', 'running shoe', 'ceramic cup', 'glass bottle', 'smartphone', 'baseball cap', 'laptop computer', 'desk lamp', 'black umbrella', 'sunglasses', 'digital camera', 'computer mouse', 'chocolate cake', 'dinner plate', 'camping tent', 'school backpack', 'candle'.
 
 \textbf{(1) CoCoCount\cite{CountGen}}. A public dataset samples object classes from COCO, which specifically includes 200 prompts with various object classes, numbers (between 2 and 10) and scenes, each prompt contains only single-class objects. 
 
 \textbf{(2) GPTSingleCount (ours)}. We build a single-class object
dataset using ChatGPT-4, which providing contextual exam-
ples for evaluation. The evaluation dataset consists of 500
prompts, with 100 prompts corresponding to each of the entity
quantities 2, 3, 5, 7, and 10. Each set of 100 prompts is further
composed of 4 prompts for each of 25 distinct object classes. Here are some examples:

\begin{itemize}
    \item \textbf{two red apples} in the basket.
    \item \textbf{Three wooden chairs} around the dining table.
    \item \textbf{Five sunflowers} in the field.
    \item \textbf{Seven office workers} in a meeting room.
    \item \textbf{Ten cars} lined up at the traffic light.
\end{itemize}
 
 \textbf{(3) GPTMultiCount (ours)}. We also build a multi-class object dataset using ChatGPT-4. This dataset consists of 100 prompts, each randomly selecting 2 or 3 object classes from a pool of 25 object classes. For each selected object class, the object quantity is assigned as 1, 2 or 3. Here are some examples:

\begin{itemize}
    \item \textbf{Three digital cameras} and \textbf{one baseball cap} on the shelf.
    \item \textbf{One school backpack} and \textbf{two smartphones} on \textbf{a wooden chair}.
    \item \textbf{One chocolate cake} on \textbf{a dinner plate} and \textbf{two glass bottles} next to it.
\end{itemize}

\subsection{Experiments Setup}
  Theoretically, the proposed CountDiffusion can be plugged into any diffusion-based T2I model. In this work, SDXL\cite{SDXL} and Pixart-$\Sigma$\cite{Pixart-sigma} are adopted as the base T2I model to demonstrate the effectiveness of the proposed CountDiffusion. SDXL has 3.3 billion parameters, using dual CLIP text encoders and an enhanced U-Net with transformer blocks for high-resolution image generation. In contrast, Pixart-$\Sigma$\ has 0.6 billion parameters, which utilizes LLM T5 as the text encoder and replaces the Unet architecture with Transformer\cite{vaswani2017attention}. Grounded SAM\cite{SAM} is leveraged for the object counting, which is one of the best segmentation models integrating object recognize and object segmentation. In our experiments, we used a total diffusion steps $T=40$ and intermediate step $t_{mid}=30$. In Grounded SAM, we set box\textunderscore threshold to 0.4, text\textunderscore threshold to 0.2, and iou\textunderscore threshold to 0.5.
  
  \textbf{Evaluation Metrics}:
  To evaluate the accurate object quantity generation ability of the proposed CountDiffusion, we report accuracy (\textbf{Acc.}) and mean absolute error (\textbf{MAE}). Acc. quantifies the percentage of correctly synthesized images that contain the same number of objects with the corresponding textual descriptions. Note that for text containing various classes of objects, the image is regarded as a correctly synthesized one if and only if all kinds of objects are generated with the correct quantities. MAE measures the difference between the generated object quantities in an image and the object quantities in its corresponding textual description. We employ Grounded SAM to obtain the object quantities required for computing Acc. and MAE. Besides, we also report the Text-to-Image Similarity using CLIP-vit-L-14 \cite{CLIP-14} (\textbf{CLIP-score}), which reflects the consistency between the generated image and the corresponding textual descriptions in the CLIP feature space. Higher CLIP-score reflects better generation results. Finally, we evaluate the human preferences in text-to-image synthesis using a widely adopted general-purpose reward model, i.e., ImageReward \cite{AestheticsScorer} (\textbf{IR}). Higher IR denotes better generation results.
  
      \begin{figure}[!b]
    \centering
    \includegraphics[width=1\linewidth]{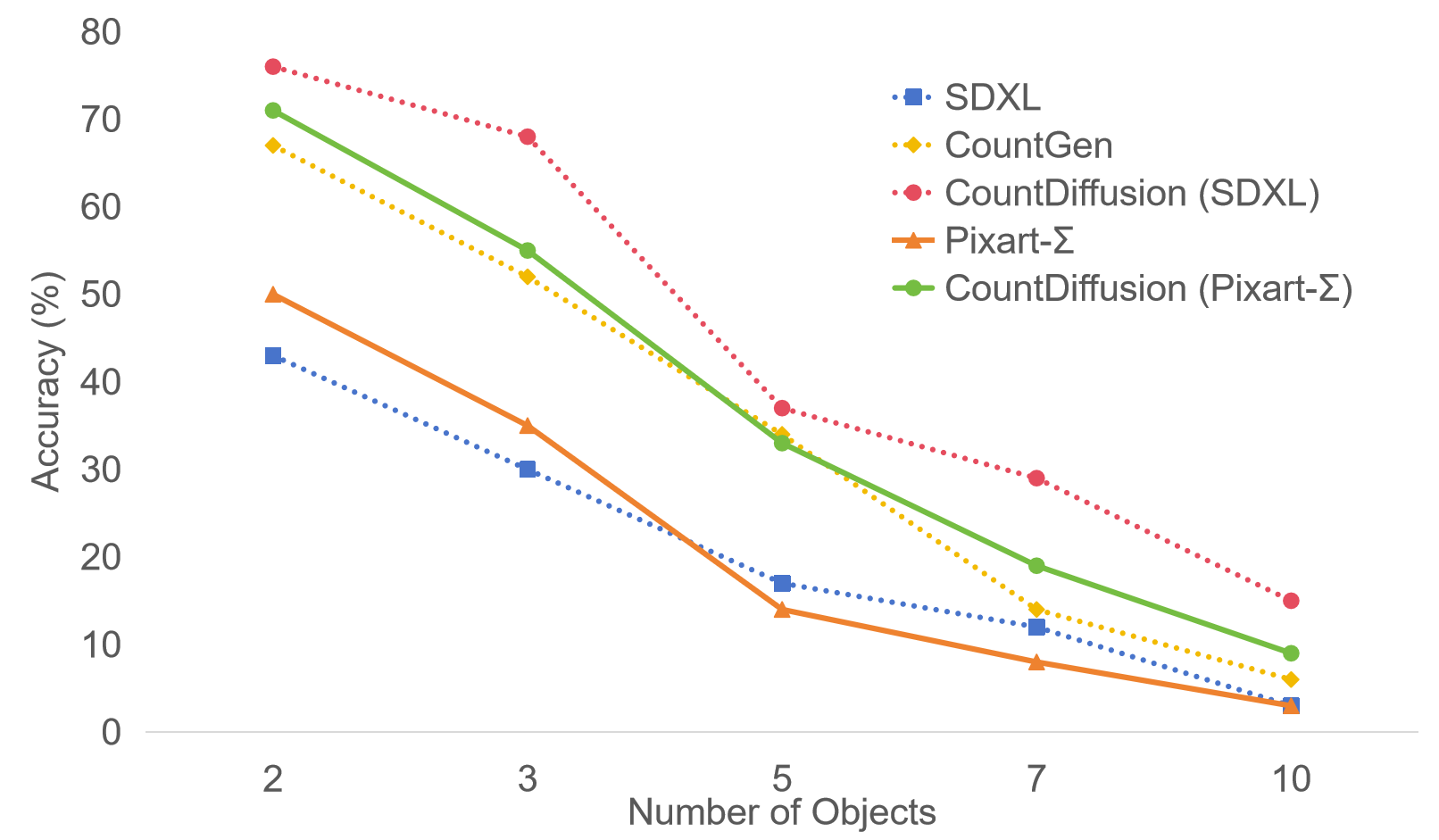}
    \caption{Comparison between the proposed CountDiffusion and state of the arts across different object quantities on GPTSingleCount dataset.}
    \label{figure4}
\end{figure} 
\subsection{Results}

    \textbf{Quantitative results.} Table \ref{table1} presents the comparison between the proposed CountDiffusion and state of the arts, where it surpasses all state-of-the-art models both on single-class and multi-class datasets in terms of accuracy, MAE and IR, demonstrating that the images generated by the proposed CountDiffusion are not only with more accurate object quantities, but also more human preferred. In most cases, CountDiffusion achieves an improvement in CLIP score, suggesting that it does not degrade the consistency between generated images and textual descriptions. Furthermore, as illustrated in Fig. \ref{figure4}, CountDiffusion consistently achieves superior performance compared to all state-of-the-art models across all object quantity scenarios.

\begin{figure}[htbp]
    \centering
    \includegraphics[width=1\linewidth]{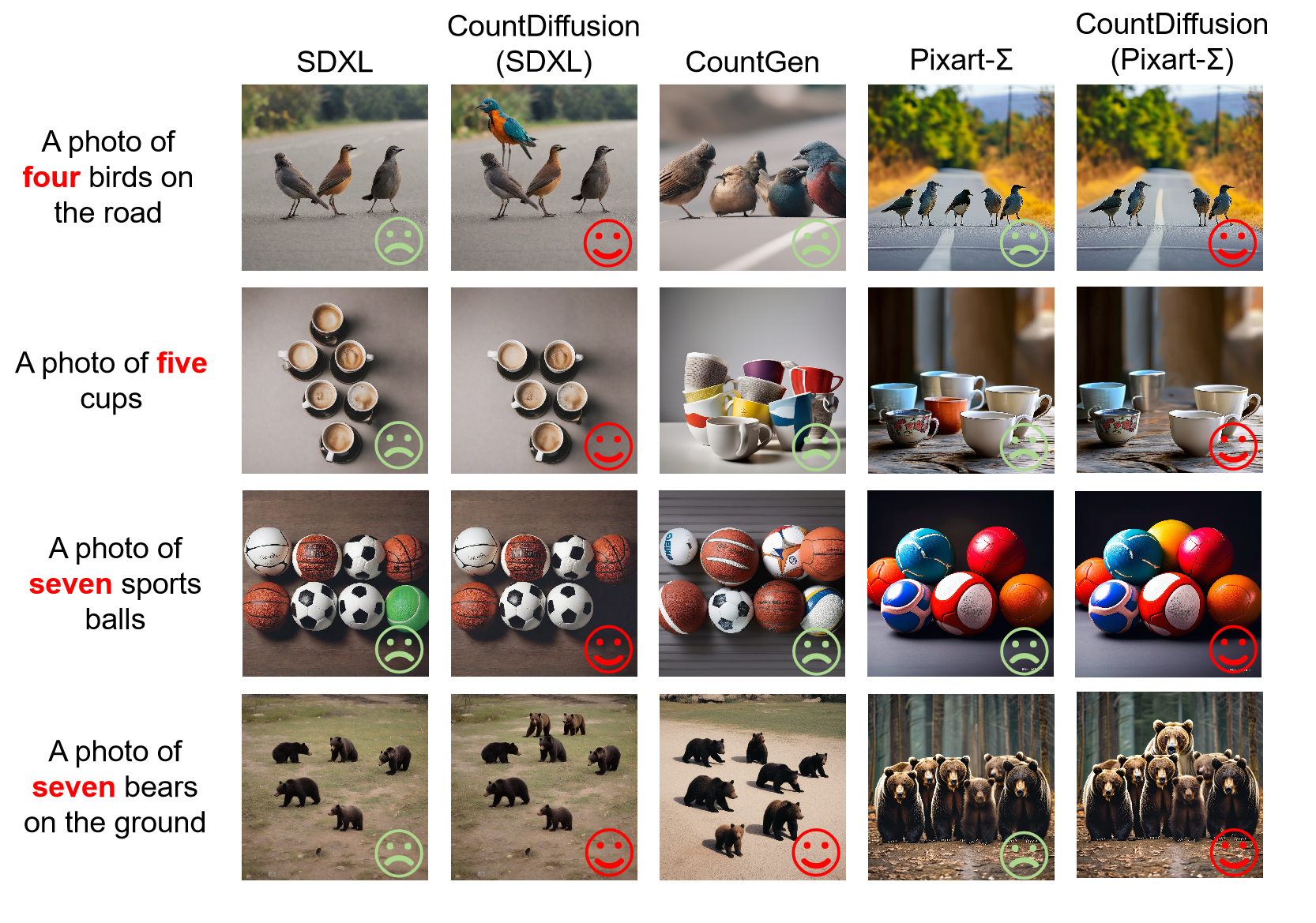}
    \caption{\textbf{Qualitative comparisons of all models.} Our method successfully generates the correct number of objects, while other methods struggle in some or all of the examples. The red smiling face in the bottom right corner of the image indicates that the correct number of objects were generated, while the green crying face indicates an error in generation.}
    \label{result}
\end{figure}

\begin{figure}[htbp]
    \centering
    \includegraphics[width=1\linewidth]{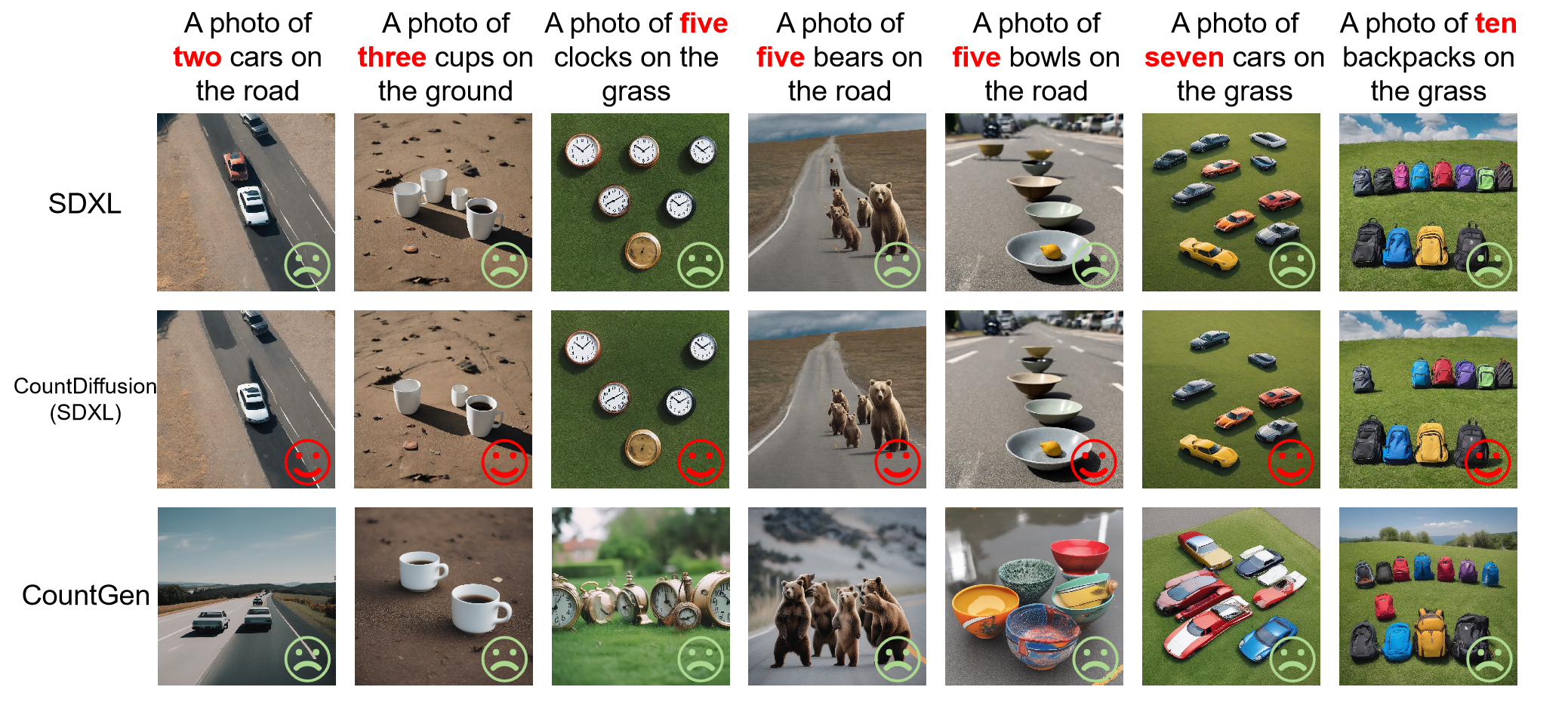}
    \caption{Qualitative comparisons based on SDXL. Our method successfully generates the correct number of objects, while SDXL struggle in all of the examples.}
    \label{SDXL}
\end{figure}

\begin{figure}[htbp]
    \centering
    \includegraphics[width=1\linewidth]{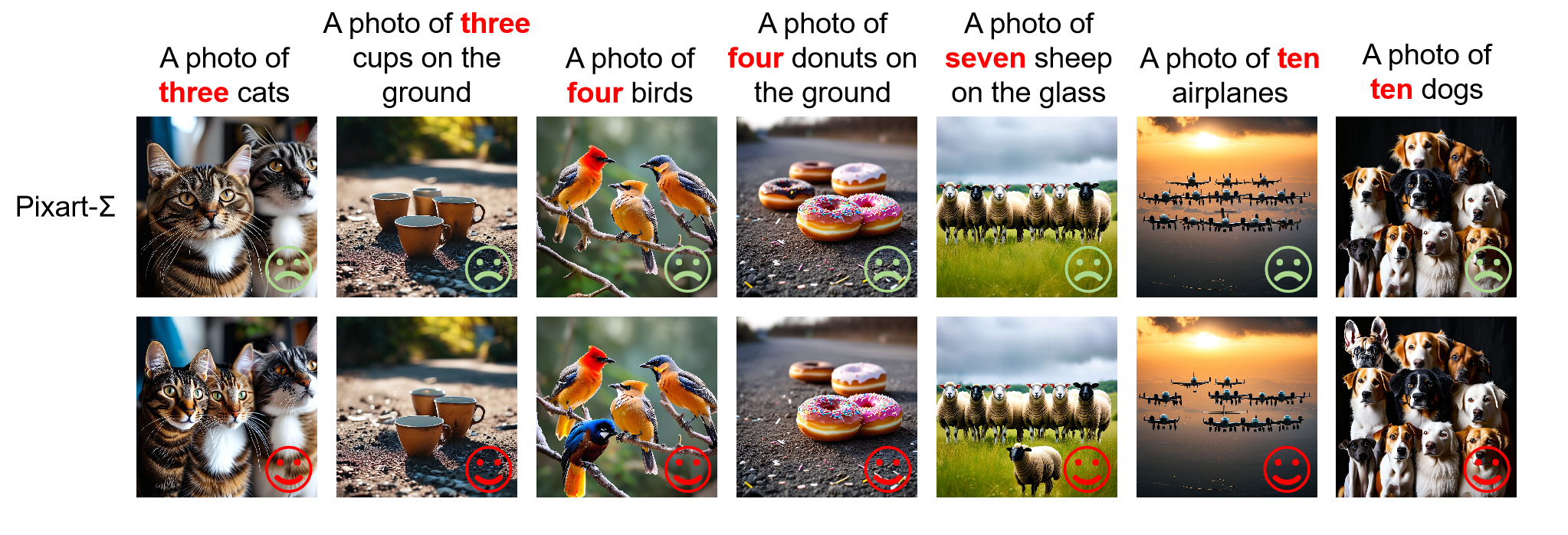}
    \caption{Qualitative comparisons based on pixart-$\Sigma$. Our method successfully generates the correct number of objects, while pixart-$\Sigma$ struggle in all of the examples.}
    \label{PIXART}
\end{figure}

\textbf{Qualitative results.} Fig. \ref{result}, Fig. \ref{SDXL}, and Fig. \ref{PIXART} show examples of prompts and the images generated by various methods. In contrast to other methods, CountDiffusion consistently generates the correct number of object instances.

   
\subsection{Ablation Study}
  Table \ref{table2} illustrates the effect of different loss value selection strategies on accuracy, MAE and CLIP-score. Specifically, \textbf{mean()} refers to averaging the values, \textbf{topk(P)} denotes selecting the top P$\%$ of values from masked attention map values, \textbf{bottomk(P)} represents selecting the bottom P$\%$ from masked attention map values, and \textbf{random(P)} means selecting a random P$\%$ of values from masked attention map values. The table shows that using mean(topk(P=50)) yields the highest accuracy. However, as P increases, excessive information may be captured, leading to unstable image generation. Conversely, decreasing P or using bottomk or random strategies results in insufficient key features, hindering accurate image correction.

  Table \ref{table3} demonstrates that as the number of universal guidance steps increases, both accuracy and CLIP score initially increase and then decrease. This can be attributed to the fact that an appropriate increase in universal guidance steps during the early stages of the generation process enhances model stability, reduces noise, and brings the image closer to the target distribution. However, further increasing the number of universal guidance steps results in insufficiently smooth mask boundaries, which in turn lowers both accuracy and CLIP scores. 
\begin{table}[htbp]
\caption{Effect of Loss Strategies on Object Quantity Generation Accuracy on CoCoCount dataset. }
\centering
\begin{tabular}{cccc}
\toprule
Loss strategies & Accuracy$\uparrow$   & MAE$\downarrow$ &   CLIP-score$\uparrow$  \\
\midrule
mean(all) & 55& 1.13& 32.1135\\
mean(topk(P=80)) & 52& 1.14& 32.1105\\
mean(topk(P=50)) & \textbf{57}& \textbf{1.02}& \textbf{32.1973}\\
mean(topk(P=10)) & 56& 1.13&32.0936\\
mean(bottomk(P=50)) & 53& 1.08& 32.0931\\
mean(random(P=50)) & 42& 1.33& 32.1469\\
\bottomrule
\end{tabular}
\label{table2}
\end{table}
\begin{table}[htbp]
\caption{Analysis of Universal Guidance Step and Total Step of Diffusion on Accuracy On CoCoCount Dataset. }
\centering
\begin{tabular}{ccccc}
\toprule
universal guidance steps & total steps & Accuracy$\uparrow$ & MAE$\downarrow$ & CLIP-score$\uparrow$        \\
\midrule
10 & 30 & 51&   1.16&32.0594\\
15 & 30 & 54&  1.14&\textbf{32.1973}\\
25 & 40 & 56&   1.01&32.0117\\
30 & 40 & 56&  0.96& 32.0154\\
35 & 40 & \textbf{58}& \textbf{0.89}& 32.0074\\
40 & 40 & 56&  1.03& 31.9234\\
\bottomrule
\end{tabular}
\label{table3}
\end{table}

\begin{figure}[htbp]
    \centering
    \includegraphics[width=1\linewidth]{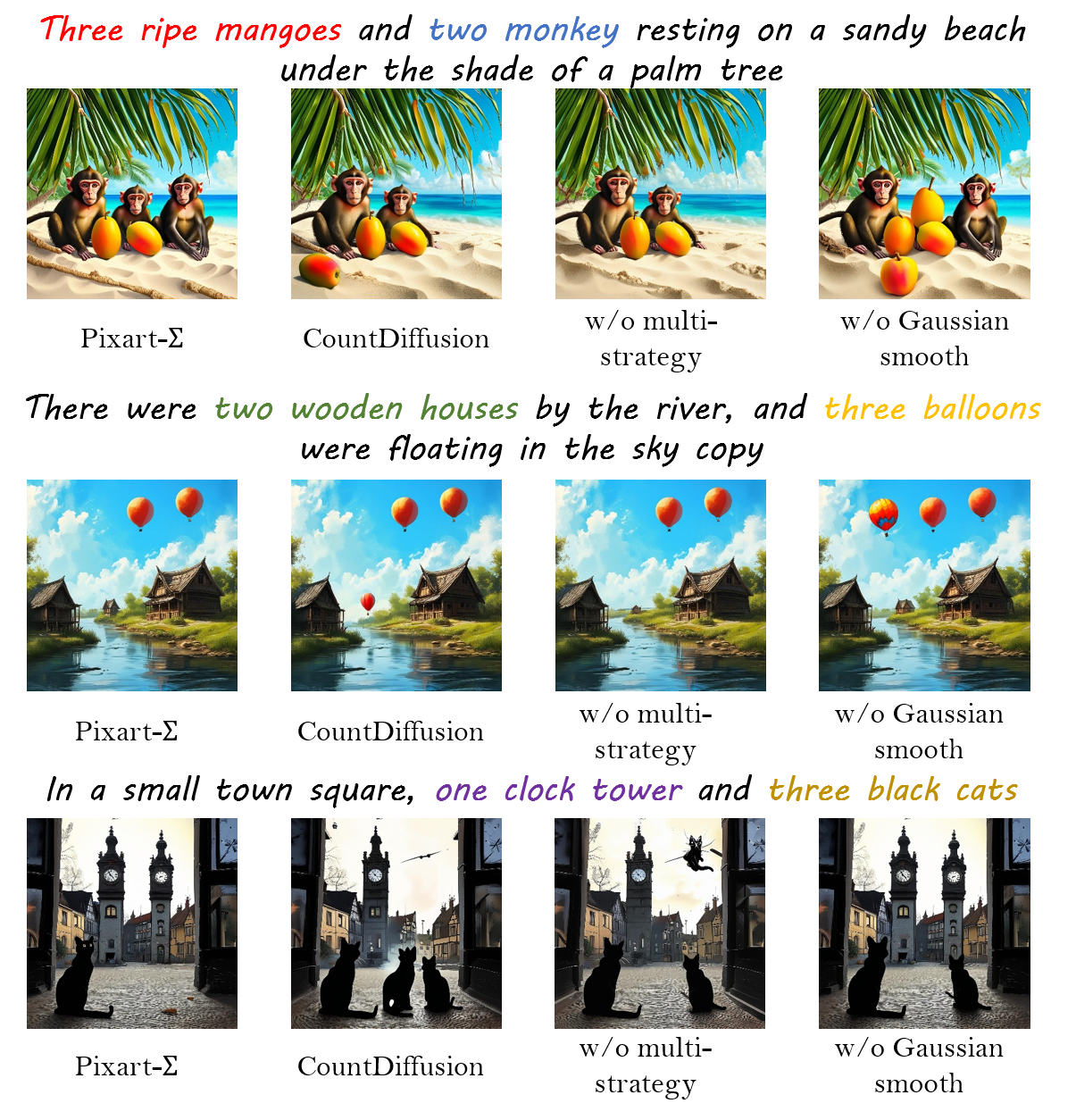}
    \caption{Ablation study of the proposed CountDiffusion.}
    \label{figure5}
\end{figure}

    We remove different components of CoundDiffusion from the original model to check their contributions to the model. As shown in Fig. \ref{figure5}, when the Multi-class Object Correction Strategy is excluded, it becomes challenging for the model to successfully generate images with correct objects counts. This is because when correcting multi-class objects simultaneously, the losses of different classes of objects will compete with each other, which makes it difficult to guide the model to correct multi-class object quantities. 

\begin{figure}[htbp]
    \centering
    \includegraphics[width=1\linewidth]{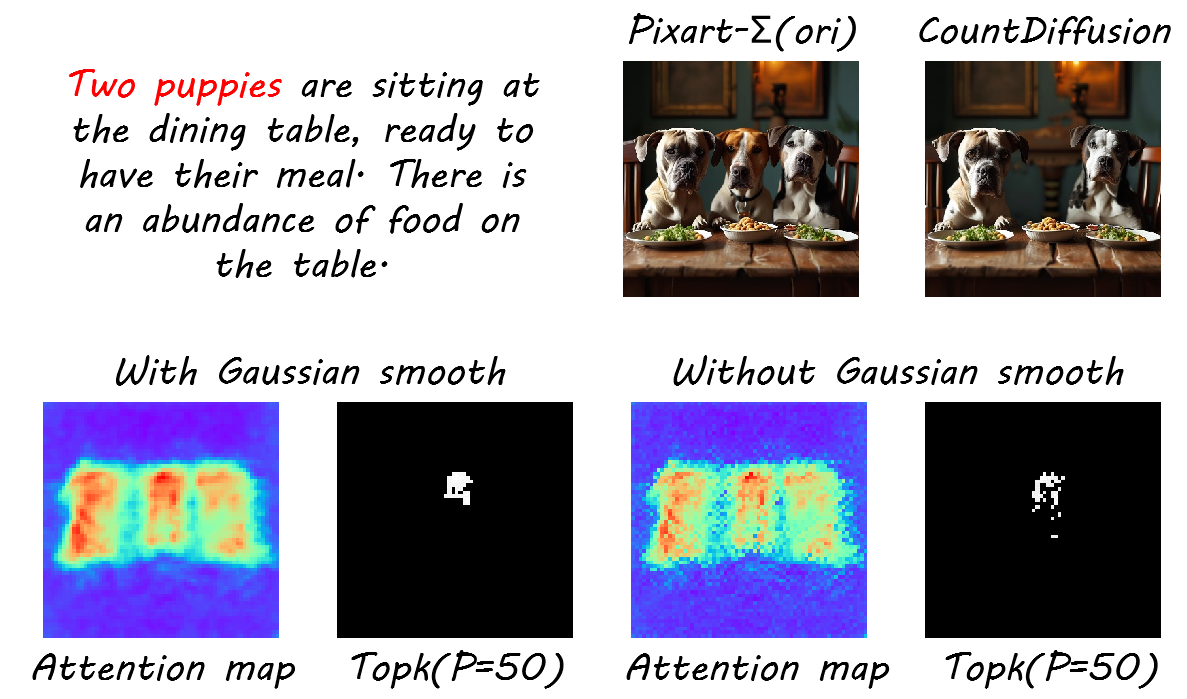}
    \caption{The effect of Gaussian smooth on attention maps. with Gaussian smooth, the region with the highest attention values tends to be more concentrated. It enables the model to successfully eliminate objects with lower control intensity.}
    \label{figure6}
\end{figure}

  Fig. \ref{figure6} illustrates the effect of Gaussian smoothing on attention maps. Calculating loss using original attention maps results in discrete high attention values ($topk(P=50)$), whereas smoothing produces a continuous region, facilitating a smoother loss decline and enhancing model performance. Moreover, Gaussian smoothing proves especially effective for object removal. It successfully remove objects at lower $\sigma$ values without degrading image quality, as excessive intensity values can lead to visual artifacts. As shown in Fig. \ref{figure6}, Gaussian smoothing directs attention towards the dog's head, and by removing the head, the entire dog is effectively removed.

\section{Conclusion and Limitation}
  This paper presents CountDiffusion, a training-free framework to improve the ability of diffusion models to synthesize images from text with correct object quantity, which consists of a detection stage to check the synthesized object quantity with an intermediate generation result and a correction stage to correct the object quantity when the generated object quantity is wrong. The model is able to seamlessly integrate with all diffusion-based T2I generation models without further training. Besides, CountDiffusion involves no human labor in the T2I synthesis process and object quantity correction stage, which makes it quite user-friendly. Experimental results demonstrate that our CountDiffusion outperforms state-of-the-art models by a huge margin. 

  It is worth noting that CountDiffusion still struggles with generating images with a large number of objects, both single-class and multi-class, limited by the inherent capabilities of the base T2I model and the counting model. We leave this as our future work.

\bibliographystyle{IEEEbib}
\bibliography{icme2025}

\end{document}